\RequirePackage{fix-cm}
\documentclass[letterpaper, 10 pt, conference]{ieeeconf}
\IEEEoverridecommandlockouts
\usepackage{graphicx}
\usepackage{float}
\usepackage{amsmath, amssymb, amsfonts}
\usepackage{cite}
\usepackage{caption}
\usepackage{subcaption}
\usepackage{hyperref}
\hypersetup{hidelinks}
\usepackage{graphics}
\usepackage{epsfig}
\usepackage{times}
\usepackage[dvipsnames]{xcolor}
\usepackage{booktabs}
\usepackage{multirow}
\usepackage{makecell}
\usepackage{pifont}
\usepackage{colortbl}
\usepackage{xspace}

\newcommand{\bench}{HM3D-M$^3$\xspace}
\newcommand{\higa}{HiGA\xspace}
\newcommand{\lavira}{LaViRA-M$^3$\xspace}
\newcommand{\dmnav}{DM$^3$-Nav\xspace}
\newcommand{\cmark}{{\color{ForestGreen}\ding{51}}}
\newcommand{\xmark}{{\color{BrickRed}\ding{55}}}
\newcommand{\nmark}{{\color{gray}--}}
\newcommand{\ie}{\emph{i.e.}\xspace}
\newcommand{\eg}{\emph{e.g.}\xspace}

\newif\iftodo
\todofalse

\newif\ifrev
\revfalse
\newcommand{\rev}[1]{\ifrev{\color{red}#1}\else{#1}\fi}

\title{\LARGE \bf
\bench: A Benchmark for Multi-Agent, Multi-Modal, \\ Multi-Object Goal Navigation
}

\author{
    \begin{tabular}{c}
        \centerline{Hongyu Ding$^{1,4,*}$, Kanyu Xu$^{2,*}$, Mingjia Chen$^{3}$, Zixuan Chen$^{2}$,} \\
        \centerline{Jieqi Shi$^{2,\dagger}$, Jing Huo$^{1,\dagger}$, Yifan Zhang$^{4,\dagger}$, Yang Gao$^{2}$}
    \end{tabular}
    \thanks{\textbf{$^{*}$Equal Contribution}, \textbf{$^{\dagger}$Corresponding Author}}
    \thanks{$^{1}$Hongyu Ding and Jing Huo are with the School of Computer Science, Nanjing University, China.}
    \thanks{$^{2}$Kanyu Xu, Zixuan Chen, Jieqi Shi and Yang Gao are with the School of Intelligence Science and Technology, Nanjing University, China.}
    \thanks{$^{3}$Mingjia Chen is with the School of Electrical and Automation Engineering, Nanjing Normal University, Nanjing, China.}
    \thanks{$^{4}$Hongyu Ding and Yifan Zhang are with the Institute of Automation, Chinese Academy of Sciences, and the University of Chinese Academy of Sciences, Nanjing, China.}
}

\begin{document}

\maketitle
\thispagestyle{empty}
\pagestyle{empty}

\begin{abstract}
Rapid progress in AI agents and robotics makes it a matter of the near future that humans are served by a team of robots.
Such a team will receive several goals at once, each specified in whatever form is convenient for the human, such as a category name, a photo, or a spoken description.
Existing benchmarks study multi-agent, multi-modal, and multi-object goal navigation in \emph{isolation}, and a benchmark that places all three in one episode is missing.
To fill this gap, we introduce \bench.
\rev{A team of $K$ agents starts at distributed locations in a photorealistic indoor scene and searches for an unordered set of $M$ object goals}, each given as an open-vocabulary category, an instance image, or an instance description.
\bench provides 1,000 validation episodes over 3,926 goals from 379 open-vocabulary categories, \rev{100K observation--assignment training samples}, and two team-level metrics, goal-level SR and makespan-normalized SPL.
We further contribute two baselines.
\higa is a trained hierarchical goal assigner on top of a frozen open-vocabulary navigation policy, and \lavira is a training-free team of VLM agents with shared mapping and \rev{progress analysis}.
Adapted multi-robot and zero-shot methods reach 16--27\% SR, while our baselines reach 38--41\% with complementary strengths and \rev{distinct failure modes}.
Furthermore, we deploy \lavira on a heterogeneous three-robot team.
Project page: \url{https://robo-lavira.github.io/hm3d-m3/}
\end{abstract}

\section{Introduction}
\label{sec:intro}

\rev{Consider a home with several robots and a resident who asks them to find a nightstand, the couch in a photo, and ``the plant by the sliding glass door''.
Nothing in this request is exotic.
Rapid progress in AI agents and robot hardware places such teams within the near future, and the request has the shape everyday instructions take: several goals at once, each given in whatever form is at hand.
What is missing is a way to measure how well a team handles it, since the goals have to be divided, recognized under three different specifications, and tracked jointly.}

\begin{figure}[t]
    \centering
    \includegraphics[width=0.97\linewidth]{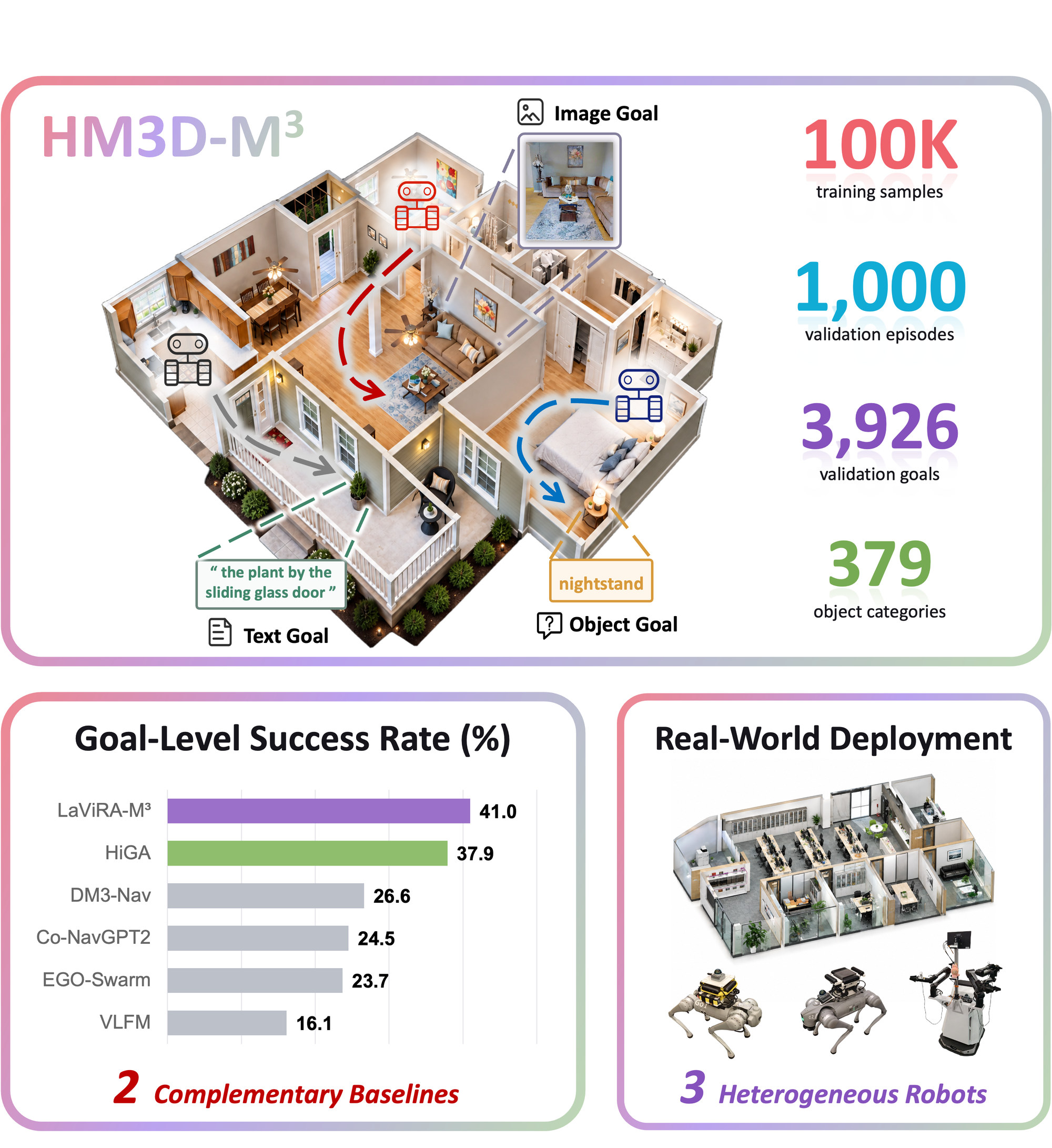}
    \vspace{2pt}
    \caption{\textbf{\bench.} \rev{Top: a team of agents searches a photorealistic scene for an unordered set of goals given as a category, an instance image, or an instance description. The benchmark provides 1,000 validation episodes, 3,926 goals, 379 categories, and 100K training samples. Bottom: goal-level SR of all methods including our two complementary baselines, and the real-world deployment on three heterogeneous robots.}}
    \label{fig:teaser}
    \vspace{-8pt}
\end{figure}
\begin{table*}[t]
\caption{Comparison with related goal-navigation benchmarks. \emph{Unord.} means that goals are given as an unordered set, and \emph{Mix} means that several modalities appear within one episode.}
\label{tab:benchmarks}
\vspace{-6pt}
\centering
\footnotesize
\begin{tabular}{l c c c c c c c c c}
\toprule
\textbf{Benchmark} & \textbf{Venue} & \textbf{Multi-agent} & \textbf{Multi-goal} & \textbf{Unord.} & \textbf{Open vocab.} & \textbf{ON} & \textbf{TN} & \textbf{IIN} & \textbf{Mix} \\
\midrule
ObjectNav~\cite{batra2020objectnav} & arXiv'20 & \xmark & \xmark & \nmark & \xmark & \cmark & \xmark & \xmark & \xmark \\
RoboTHOR~\cite{deitke2020robothor} & CVPR'20 & \xmark & \xmark & \nmark & \xmark & \cmark & \xmark & \xmark & \xmark \\
MultiON~\cite{wani2020multion} & NeurIPS'20 & \xmark & \cmark & \xmark & \xmark & \cmark & \xmark & \xmark & \xmark \\
CollaVN~\cite{wang2021collavn} & NeurIPS'21 & \cmark & \xmark & \nmark & \xmark & \xmark & \xmark & \xmark & \xmark \\
InstanceImageNav~\cite{krantz2022iin} & arXiv'22 & \xmark & \xmark & \nmark & \xmark & \xmark & \xmark & \cmark & \xmark \\
HM3D-OVON~\cite{yokoyama2024ovon} & IROS'24 & \xmark & \xmark & \nmark & \cmark & \cmark & \xmark & \xmark & \xmark \\
GOAT-Bench~\cite{khanna2024goatbench} & CVPR'24 & \xmark & \cmark & \xmark & \cmark & \cmark & \cmark & \cmark & \cmark \\
Habitat-MAS~\cite{chen2025emos} & ICLR'25 & \cmark & \cmark & \cmark & \xmark & \cmark & \xmark & \xmark & \xmark \\
HM3D-MFMON~\cite{hm3dmfmon2026} & arXiv'26 & \xmark & \cmark & \xmark & \xmark & \cmark & \xmark & \xmark & \xmark \\
\midrule
\textbf{\bench (ours)} & \nmark & \cmark & \cmark & \cmark & \cmark & \cmark & \cmark & \cmark & \cmark \\
\bottomrule
\end{tabular}
\vspace{-6pt}
\end{table*}

Standing in that future and looking back at the present, the landscape of navigation benchmarks is fragmented, as Table~\ref{tab:benchmarks} summarizes.
Single-agent object-goal navigation asks for one instance of one category, recently from an open vocabulary~\cite{batra2020objectnav,yokoyama2024ovon}. Multi-object variants give one agent several category goals~\cite{wani2020multion,gireesh2023seqagnostic}. Multi-modal variants mix category, image, and language goals, but present them sequentially and score each sub-task on its own~\cite{khanna2024goatbench}.
Multi-robot semantic navigation evaluates teams, but on a single goal of a single modality, and grants success as soon as \emph{any} robot arrives~\cite{yu2023conavgpt,shen2025mcoconav}.
Recent systems combine all three~\cite{kashiri2026dm3nav,anygoal2026}, yet they are evaluated on episodes borrowed from single-agent benchmarks with per-sub-task scoring.
No benchmark defines the joint task, generates episodes for it, or measures the team as a whole.

The intersection is not a concatenation of its parts.
Allocation happens under partial observability, since who should pursue which goal depends on heterogeneous evidence that arrives as the map grows. Verification becomes instance-level, since an image or a description refers to one specific object. Completion is joint, since the natural cost is the team makespan, so an idle or redundant robot is not free.

To fill this gap, we present \bench, a benchmark for \textbf{M}ulti-agent, \textbf{M}ulti-modal, \textbf{M}ulti-object goal navigation in photorealistic scans of real buildings~\cite{ramakrishnan2021hm3d,yadav2023hm3dsem}, illustrated in Fig.~\ref{fig:teaser}.
\rev{An episode places $K$ agents at distributed, jointly reachable start positions and gives the team an unordered set of $M$ goals from 379 open-vocabulary categories.
Each goal is specified as a category (ON), an instance image (IIN), or an instance description (TN).
A goal counts as found when an agent declares it while standing within 1\,m of that instance, and every episode stores the oracle team makespan $d^\ast$ for budgeting and normalization.}

We further contribute two baselines, one trained and one training-free.
\higa, short for Hierarchical Goal Assigner, is a \emph{trained} method in which a transformer reads the shared map, agent poses, and per-goal CLIP heat maps, assigns goals to agents, and hands each assignment to a frozen open-vocabulary navigation policy~\cite{yokoyama2024ovon}. It is trained with behavior cloning and DAgger~\cite{ross2011dagger} from an oracle assigner.
\lavira is a \emph{training-free} method that extends the vision-language model (VLM) navigation agent of LaViRA~\cite{ding2025lavira,ding2026unilavira} to a team coordinated only through a shared map, a shared goal board, and \rev{shared natural-language progress analysis}.
We also adapt four existing multi-robot and zero-shot methods~\cite{yokoyama2024vlfm,zhou2021egoswarm,yu2023conavgpt,kashiri2026dm3nav} to the benchmark.

The results expose a task that is far from solved, with adapted methods at 16--27\% SR, \higa at 37.9\%, and \lavira at 41.0\%, and two baselines with complementary strengths.
\higa is the strongest on category goals but weak on image goals and inefficient, whereas \lavira is the strongest on image and text goals and the most efficient.
\rev{Their failures differ, since \lavira mostly declares a goal from the wrong place, whereas \higa mostly fails to detect or match it.}

Our contributions are:
\begin{itemize}
    \item \rev{\bench, the first benchmark for multi-agent, multi-modal, multi-object goal navigation, with a standardized episode generator, a validation set of 1,000 episodes and 3,926 goals on 36 held-out scenes, a training set of 100K observation--assignment samples on disjoint scenes, and team-level metrics.}
    \item \rev{Two baselines, the trained \higa and the training-free \lavira, plus adaptations of four existing methods, evaluated with ablations, scaling studies over $K$ and $M$, and a per-goal failure attribution.}
    \item \rev{A real-world deployment of \lavira on a heterogeneous team of three robots running local VLMs, which complete mixed-modality goal sets in an office through the same coordination channels as in simulation.}
\end{itemize}

\section{Related Work}
\label{sec:related}

\begin{figure*}[t]
    \centering
    \includegraphics[width=0.97\linewidth]{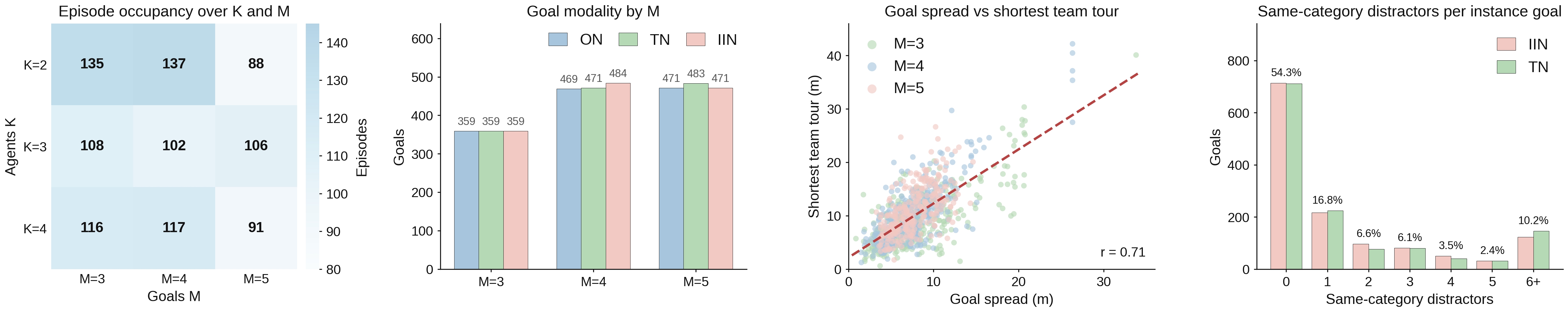}
    \vspace{-2pt}
    \caption{\textbf{\bench validation set with 1,000 episodes and 3,926 goals.} (a) Episodes per $(K, M)$ cell. (b) Goals per modality for each $M$. (c) Goal spread, the mean pairwise geodesic distance between goals, against the shortest team tour $d^\ast$, correlated with $r{=}0.71$. \rev{(d) Number of same-category instances on the same floor for each IIN and TN goal.}}
    \label{fig:dataset}
    \vspace{-8pt}
\end{figure*}

\noindent\textbf{Goal navigation benchmarks.}
Single-agent benchmarks have progressively enriched how a goal is specified, from a closed set of categories~\cite{batra2020objectnav,deitke2020robothor} to an open vocabulary~\cite{yokoyama2024ovon}, and to a particular instance shown in an image~\cite{krantz2022iin} or described in language~\cite{khanna2024goatbench}.
Multi-object variants ask one agent for several goals, in a fixed order~\cite{wani2020multion} or as an unordered set~\cite{gireesh2023seqagnostic}, \rev{recently across several floors~\cite{hm3dmfmon2026}}, and lifelong multi-modal navigation chains 5--10 goals of mixed modality into one episode with per-sub-task scoring~\cite{chang2023goat,khanna2024goatbench}.
\rev{The one multi-agent benchmark among them, Habitat-MAS~\cite{chen2025emos}, targets language-specified household tasks for heterogeneous teams, with multi-object navigation limited to a closed vocabulary and a single modality.}
\bench inherits the open vocabulary and the three goal modalities of this line, and adds a team of agents, an unordered goal set \rev{searched jointly}, and a per-episode team optimum.

\noindent\textbf{Multi-robot semantic navigation.}
After early work in synthetic scenes~\cite{wang2021collavn}, recent systems on photorealistic scans let a VLM allocate frontiers to robots~\cite{yu2023conavgpt,shen2025mcoconav}, and LLM agents have been coordinated on unordered multi-object missions in a single modality on small custom episode sets~\cite{wu2024camon}.
\rev{Closest to our task are two very recent systems that run robot teams on multi-modal, multi-goal missions.
\dmnav~\cite{kashiri2026dm3nav} is decentralized, with agents that exchange local maps, task states, and intents within communication range and claim goals by broadcasting them, and it introduces a makespan-normalized SPL.
AnyGoal~\cite{anygoal2026} is a training-free team in which a VLM guides exploration toward category, image, and language goals presented sequentially, as in GOAT-Bench.
Both are methods rather than benchmarks, evaluated on single-agent episodes with sub-task scoring, and neither defines a multi-agent episode distribution, a team budget, or team metrics, which \bench supplies.
We evaluate the official implementation of \dmnav on it and adopt its MSPL. The code of AnyGoal had not been released at the time of writing.}

\noindent\textbf{\rev{Trained and training-free navigation policies.}}
Trained object-goal policies range from end-to-end RL and imitation~\cite{ramrakhya2023pirlnav} to the DAgger-trained open-vocabulary policy~\cite{yokoyama2024ovon} that we use frozen as the lower level of \higa.
Training-free methods pair explicit maps with foundation models~\cite{gadre2023cow,yokoyama2024vlfm}, unify category, image, and text goals by graph matching~\cite{yin2025unigoal}, or let an LLM or VLM act as the navigation agent itself~\cite{long2024instructnav,ding2025lavira}.
LaViRA~\cite{ding2025lavira} and Uni-LaViRA~\cite{ding2026unilavira} decompose each decision into a language, a vision, and a robot action. \lavira extends them to a team.

\section{The \bench Benchmark}
\label{sec:benchmark}

\subsection{Task Definition}
\label{sec:task}
An episode of \bench is a tuple $(S, \{s_k\}_{k=1}^{K}, \{g_j\}_{j=1}^{M})$ with a scene $S$, start poses $s_k$ for $K$ agents, and $M$ goals.
Each goal $g_j=(o_j, m_j, x_j)$ refers to an object instance $o_j$ of a distinct category and is presented to the team in one modality $m_j$ with content $x_j$.
An \textbf{ON} goal gives the category name, \eg ``nightstand''.
An \textbf{IIN} goal gives an image of the specific instance taken from a camera that is not the agent's, and a \textbf{TN} goal gives a description of the specific instance, \eg ``the plant by the sliding glass door''.
\rev{IIN and TN goals point to a specific instance, while an ON goal accepts any instance of the named category.}
All $M$ goals are given at the start in no particular order.
Each agent observes posed RGB-D images and its own pose in a frame shared by the team, as in prior multi-robot work~\cite{yu2023conavgpt,kashiri2026dm3nav}, and executes discrete actions, \texttt{MOVE\_FORWARD} by 0.25\,m and \texttt{TURN\_LEFT} or \texttt{TURN\_RIGHT} by $30^\circ$, in lockstep with its teammates.

\noindent\textbf{Success protocol.}
At any step an agent may declare $\texttt{FOUND}(j)$ for a goal $j$ that is still unresolved.
The declaration succeeds if the agent stands within 1\,m of a navigable viewpoint of an acceptable instance of $g_j$, following the success criterion of GOAT-Bench~\cite{khanna2024goatbench}, and is otherwise void, with the episode continuing without penalty.
\rev{Declaring and scoring are separate: a declaration is the policy's own belief, which the team may share, whereas the evaluator scores it against the ground truth offline and never reveals the verdict, so any verification before declaring must use the agent's own perception.}
A goal counts for the team regardless of which agent found it.
The episode ends when all $M$ goals are found or when every agent has exhausted its budget of 500 actions or a path length of $3\,d^\ast$, where $d^\ast$ is the oracle team makespan defined below.

\subsection{Episode Generation}
\label{sec:generation}
\noindent\textbf{Scenes and goals.}
We use the HM3DSem scenes~\cite{yadav2023hm3dsem} with the train/validation scene split of HM3D-OVON~\cite{yokoyama2024ovon}, and draw goal instances and their navigable viewpoints from its 379 open-vocabulary categories.
For each episode we sample a floor that contains at least $M$ distinct categories, then $M$ instances of $M$ distinct categories on that floor, each keeping up to eight navigable viewpoints as its success region.
Modalities are balanced within the episode, with exactly one goal per modality for $M{=}3$ and as evenly as possible otherwise, as shown in Fig.~\ref{fig:dataset}b.
\rev{Fig.~\ref{fig:dataset}d shows how many other instances of the same category share the floor with each IIN and TN goal.}

\noindent\textbf{Image and text goals.}
IIN images follow the InstanceImageNav protocol~\cite{krantz2022iin} adopted by GOAT-Bench, with a camera placed 0.5--2\,m from the instance at a height of 0.8--1.5\,m and a field of view of $60$--$120^\circ$, keeping only viewpoints from which the instance is sufficiently visible.
For TN descriptions we follow the LanguageNav pipeline of GOAT-Bench, rendering the instance from its best viewpoint and prompting a VLM, Qwen3.5-27B~\cite{qwen35}, for one sentence that distinguishes it from others of its category by color, material, and neighboring objects, and we spot-check the results.

\noindent\textbf{Starts and reachability.}
The $K$ start poses are sampled on the same floor by rejection sampling.
They must lie on navigable regions of sufficient size, be separated by a controlled geodesic distance so that the team starts distributed rather than co-located, and jointly cover the goals, \ie every goal must be geodesically reachable from at least one start.
\rev{Each goal enters the routing geometry through a single representative viewpoint, and the start--goal and goal--goal geodesic matrices over these points are stored with the episode.}

\noindent\textbf{Oracle makespan.}
From these matrices we compute the oracle team makespan
\begin{equation}
d^\ast = \min_{a:\{1..M\}\to\{1..K\}} \; \max_{k} \; \ell_k\big(a^{-1}(k)\big),
\label{eq:dstar}
\end{equation}
where $a$ assigns goals to agents and $\ell_k(\cdot)$ is the length of agent $k$'s open nearest-neighbor tour from $s_k$ through its assigned goals.
The minimization is exact by enumeration when $K^M\le 10^5$ and uses a load-balanced greedy assignment otherwise.
\rev{$d^\ast$ is thus the optimal division of labor under nearest-neighbor routing rather than a global optimum, and since the per-agent tour is not always the shortest one and ON goals use the sampled instance although any instance is acceptable, it is an upper bound on the true optimum. It scales the budget and normalizes efficiency for all methods alike.}

\noindent\textbf{Splits.}
The validation set contains 1,000 episodes on \rev{the 36 HM3D-OVON validation scenes}, sampled uniformly over $K\in\{2,3,4\}$ and $M\in\{3,4,5\}$ and stratified over scenes, as shown in Fig.~\ref{fig:dataset}a, with 3,926 goals of which 1,299 are ON, 1,313 are TN, and 1,314 are IIN.
\rev{The training set is generated on the disjoint HM3D-OVON training scenes with the same procedure. From it we derive 100K observation--assignment samples for learning coordination, as described in Sec.~\ref{sec:higa}.}

\subsection{Metrics}
\label{sec:metrics}
Let $c_i\le M_i$ be the number of goals found in episode $i$ and $T_i=\max_k \ell_k$ the team makespan, \ie the path length of the agent that travelled farthest.
Goal-level \textbf{SR} is $\frac{1}{N}\sum_i c_i/M_i$, \ie the fraction of goals the team finds, which equals the \emph{Progress} of MultiON~\cite{wani2020multion} and the sub-task success of GOAT-Bench.
\rev{We report goal-level rather than episode-level success because an episode succeeds only if all $M$ goals are found, which current methods achieve rarely. An episode-level rate would sit near zero and hide partial progress and per-modality differences.}
Makespan-normalized \textbf{SPL} is $\frac{1}{N}\sum_i \frac{c_i}{M_i}\cdot\frac{d^\ast_i}{\max(d^\ast_i, T_i)}$, the multi-agent SPL of~\cite{kashiri2026dm3nav}.
Normalizing by the makespan rather than the summed path rewards division of labor.
\rev{Because SPL is weighted by $c_i/M_i$, a team that finds only its nearest goals and stops early also scores low. In-place rotations add no path length but consume the action budget.}
\textbf{ON}, \textbf{TN}, and \textbf{IIN} are the found rates of goals of each modality.

\section{Baselines}
\label{sec:baselines}

\begin{figure*}[t]
    \centering
    \captionsetup[subfigure]{justification=raggedright,singlelinecheck=false}
    \begin{subfigure}{\linewidth}
        \raggedright
        \includegraphics[width=0.9\linewidth]{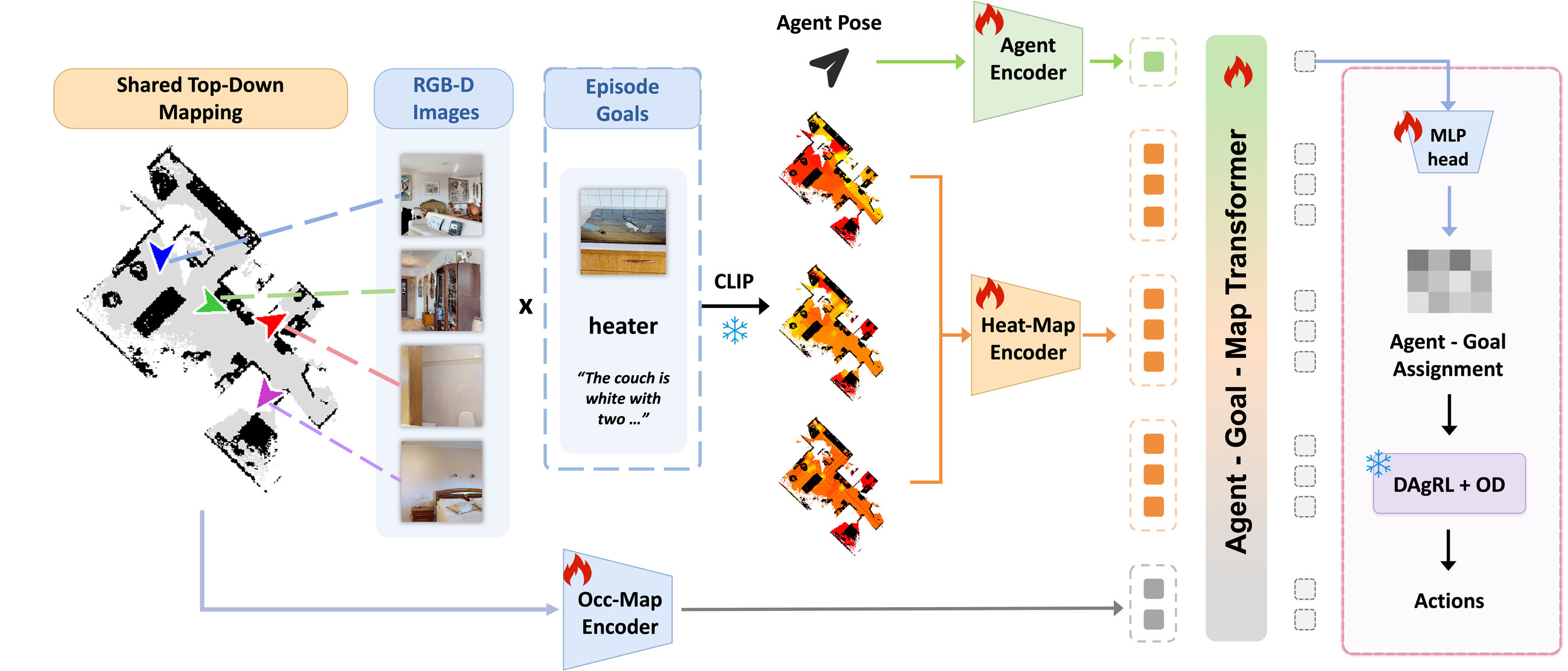}
        \caption{\higa, trained. A transformer over agent, goal, and map tokens assigns goals to agents, and each agent executes its assignment with a frozen open-vocabulary navigation policy. \rev{Flames mark trained modules and snowflakes frozen ones.}}
        \label{fig:higa}
    \end{subfigure}\par\vspace{2pt}
    \begin{subfigure}{\linewidth}
        \centering
        \includegraphics[width=0.9\linewidth]{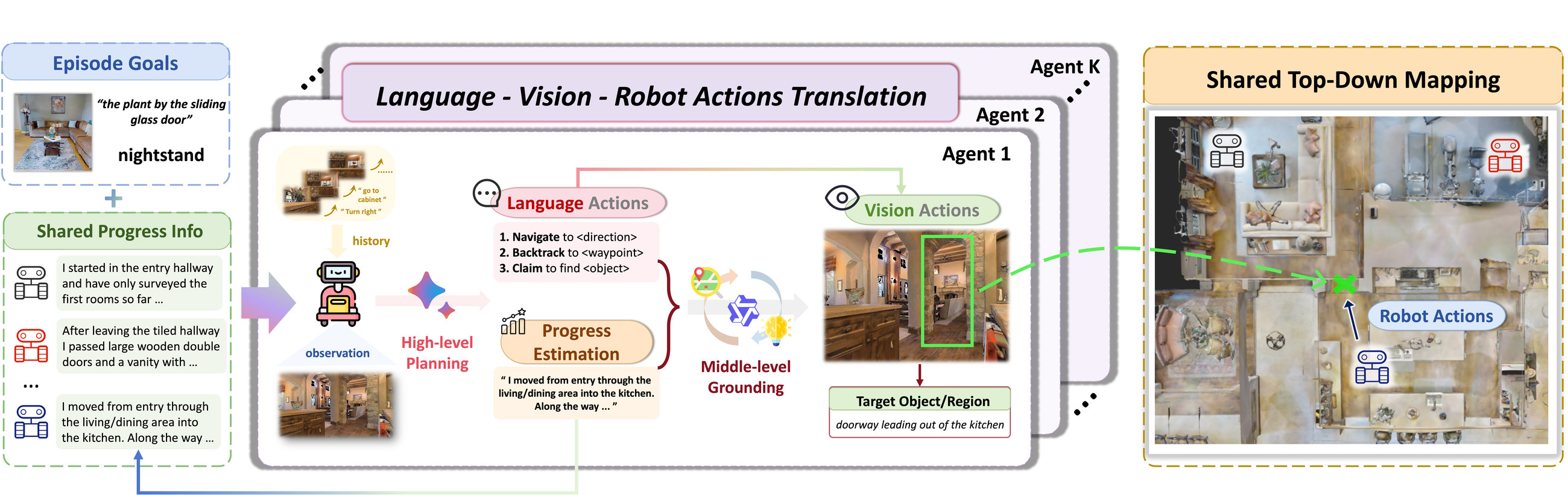}
        \caption{\lavira, training-free. Every agent runs a language--vision--robot action agent, and the team shares a top-down map, a goal board, and \rev{natural-language progress analysis}. No goal is ever explicitly assigned.}
        \label{fig:lavira}
    \end{subfigure}\par\vspace{-4pt}
    \caption{\textbf{The two \bench baselines.}}
    \label{fig:pipelines}
    \vspace{-8pt}
\end{figure*}

All methods receive the same inputs, the goal set with category names, reference images, and descriptions, posed RGB-D from every agent, and a shared top-down \rev{occupancy} map built from the team's depth observations.
\rev{Goals that any agent has declared found are removed from every agent's list. The evaluator's verdict on those declarations is never revealed to the team.}

\subsection{Adapted Methods}
\label{sec:adapted}
To compare \emph{coordination} strategies rather than perception stacks, all adapted methods and \higa share one goal-recognition module, taken from the strongest HM3D-OVON baseline~\cite{yokoyama2024ovon}.
An open-vocabulary detector, OWLv2~\cite{minderer2023owlv2}, proposes candidates for every unresolved goal in every frame.
A candidate is accepted for an ON goal by detector confidence, for a TN goal by CLIP~\cite{radford2021clip} similarity between the crop and the description, and for an IIN goal by SuperGlue~\cite{sarlin2020superglue} keypoint matching against the reference image, after which a point-goal policy approaches the object and declares \texttt{FOUND}.
Exploration and coordination follow each method's original design.
\textbf{VLFM}~\cite{yokoyama2024vlfm} scores frontiers with a BLIP-2~\cite{li2023blip2} value map, and each agent greedily heads to the highest-value frontier for the unresolved goals on the shared map.
\textbf{EGO-Swarm}~\cite{zhou2021egoswarm} is a purely geometric swarm in which each agent moves to the nearest frontier, with a repulsion term that keeps agents apart.
\textbf{Co-NavGPT2}~\cite{yu2023conavgpt} uses a centralized VLM planner that reads the shared semantic top-down map with agent positions and frontier candidates and assigns one frontier to every agent at each replanning step. \rev{We use Qwen3.5-27B in place of GPT-4o so that all VLM-based methods share the same model.}
\textbf{\dmnav}~\cite{kashiri2026dm3nav} is decentralized, with agents exchanging local maps, task states, and navigation intents within communication range, claiming goals by broadcasting their intent, and selecting frontiers weighted by distance. \rev{We port the official implementation to our episodes.}

\subsection{\higa, a Hierarchical Goal Assigner}
\label{sec:higa}
\higa splits the problem into two levels, shown in Fig.~\ref{fig:higa}.
An upper level decides \emph{who pursues what}, and a lower level decides \emph{how to get there}.

\noindent\textbf{Lower level.}
Each agent executes its current assignment with the frozen DAgRL+OD policy of HM3D-OVON~\cite{yokoyama2024ovon}, whose RGB policy explores toward the assigned category until the recognition module of Sec.~\ref{sec:adapted} takes over.
The detector watches all unresolved goals rather than only the assigned one, so an agent that passes an unassigned goal will still claim it.

\noindent\textbf{Upper level.}
The assigner observes the occupancy map, the poses of the $K$ agents, and one heat map per goal, all in the shared frame.
The heat map is obtained by scoring every RGB frame of every agent against the goal text with CLIP~\cite{radford2021clip} and back-projecting the score through depth onto the map, where image goals use their category name.
Map, goal, and agent tokens from three encoders are processed jointly by a 3-layer transformer with $d{=}256$, an MLP head scores every agent--goal pair, and the Hungarian algorithm decodes a one-to-one assignment, leaving surplus goals unassigned and surplus agents exploring.
The assignment is recomputed whenever a goal is found, an agent becomes idle, or a new candidate is detected.

\noindent\textbf{Training.}
The teacher is an oracle assigner that uses ground-truth goal positions to compute the makespan-optimal assignment of Eq.~\eqref{eq:dstar} at every event.
We collect 100K observation--assignment samples from the teacher on the training scenes, train with cross-entropy to the teacher's assignment as behavior cloning, and then run three rounds of DAgger~\cite{ross2011dagger}, relabeling the states visited by the student.

\subsection{\lavira, a Team of VLM Agents}
\label{sec:lavira}
Each agent is an instance of the LaViRA agent~\cite{ding2025lavira,ding2026unilavira}, shown in Fig.~\ref{fig:lavira}, which translates a decision through three levels.
\rev{In the \emph{language action}, a VLM, Qwen3.5-27B, reads the goal set including reference images and descriptions, a four-view panorama, and its navigation history.
It outputs a \emph{progress estimation} together with a high-level action, which navigates in a direction, backtracks to a waypoint, or claims to have found a goal.}
In the \emph{vision action}, a VLM grounds the chosen direction to a bounding box, and in the \emph{robot action}, the agent plans to the grounded target with the fast marching method on the shared map.

\noindent\textbf{Team coordination.}
Three channels connect the agents.
The \emph{shared top-down map}, \rev{a geometric occupancy map}, is used for planning and to avoid re-exploring regions a teammate has covered, and it is not shown to the VLM.
The \emph{shared goal board}, SG, lists every goal with its status, either unresolved or found by a given agent.
\rev{\emph{Shared progress analysis}, SP,} prepends each agent's latest progress estimation, a first-person summary of where it has been and what it has seen, to every teammate's prompt.
Division of labor is not enforced and emerges from the VLM reading the board and its teammates' \rev{progress analysis}.

\noindent\textbf{Declaring a goal.}
When the language action \rev{proposes to claim goal $j$}, the agent re-examines the object in a four-view panorama with the VLM, requires the object to be within 2\,m in depth, and requires the VLM's identity judgment against the goal specification to be confident.
Only then is \texttt{FOUND}$(j)$ declared and written to the board.
\rev{The 2\,m gate is a property of this baseline and is looser than the 1\,m success radius. Sec.~\ref{sec:failure} quantifies its share of the failures.}

\section{Experiments}
\label{sec:experiments}

\subsection{Setup}
All experiments run in Habitat~\cite{savva2019habitat} on the 1,000 validation episodes of \bench with a single evaluation run per method on 8 NVIDIA A800 GPUs.
Agents act in lockstep with the discrete action space of Sec.~\ref{sec:task}, and each agent has a budget of 500 actions and a path length of $3\,d^\ast$.
\higa and the adapted methods observe through the Stretch camera of their base policy, with $360{\times}640$ RGB-D and a $42^\circ$ field of view, and \lavira observes $640{\times}480$ RGB-D with a $79^\circ$ field of view.
\rev{Since the field of view differs, we additionally evaluate \lavira with the same $42^\circ$ Stretch camera as a sensor-matched control in Table~\ref{tab:ablation}.}
Co-NavGPT2, \lavira, and the TN descriptions all use Qwen3.5-27B~\cite{qwen35}, and \higa and the adapted methods use the detector-based recognition module of Sec.~\ref{sec:adapted}.
All numbers are in percent.

\begin{table*}[t]
\caption{\textbf{Main results} on the \bench validation split with 1,000 episodes. SR and SPL are the goal-level SR and the makespan-normalized SPL of Sec.~\ref{sec:metrics}, and ON, TN, and IIN are found rates per modality. Best in \textbf{bold}.}
\label{tab:main}
\vspace{-6pt}
\centering
\small
\renewcommand{\arraystretch}{1.1}
\setlength{\tabcolsep}{11pt}
\begin{tabular}{l l c c c c c}
\toprule
\textbf{Method} & \textbf{Coordination} & \textbf{SR}$\uparrow$ & \textbf{SPL}$\uparrow$ & \textbf{ON}$\uparrow$ & \textbf{TN}$\uparrow$ & \textbf{IIN}$\uparrow$ \\
\midrule
VLFM~\cite{yokoyama2024vlfm} & independent agents on a shared map & 16.1 & 11.8 & 24.5 & 17.7 & 5.9 \\
EGO-Swarm~\cite{zhou2021egoswarm} & geometric dispersion & 23.7 & 10.7 & 35.5 & 28.0 & 7.5 \\
Co-NavGPT2~\cite{yu2023conavgpt} & centralized VLM frontier allocation & 24.5 & 9.4 & 37.3 & 28.8 & 7.6 \\
\dmnav~\cite{kashiri2026dm3nav} & decentralized intent broadcasting & 26.6 & 9.9 & 35.9 & 30.4 & 14.2 \\
\midrule
\textbf{\higa} (ours) & learned goal assignment & 37.9 & 7.1 & \textbf{53.0} & 40.2 & 18.8 \\
\textbf{\lavira} (ours) & shared goal board and \rev{progress analysis} & \textbf{41.0} & \textbf{20.3} & 38.6 & \textbf{40.9} & \textbf{41.4} \\
\bottomrule
\end{tabular}
\renewcommand{\arraystretch}{1.0}
\vspace{-8pt}
\end{table*}

\subsection{Main Results}
Table~\ref{tab:main} compares the four adapted methods with our two baselines, and three observations stand out.

\noindent\textbf{The task is far from solved.}
Adapted methods find 16--27\% of the goals, the best method, \lavira, finds 41.0\%, and the best SPL is 20.3\%.
Image goals are the hardest for every detector-based method, with 5.9--18.8\%.

\noindent\textbf{Frontier allocation is not goal allocation.}
Among the adapted methods, spreading the team out matters more than semantic frontier scoring.
The purely geometric EGO-Swarm at 23.7\% outperforms independent VLFM agents at 16.1\%, and the VLM- and intent-based frontier allocators of Co-NavGPT2 and \dmnav add only a few points, 24.5\% and 26.6\%.
\rev{Once the map is shared, deciding \emph{which goal} each agent pursues is what separates the methods.
\higa, which shares the camera and the recognition module of the adapted methods, gains 11 points over the best of them.
\lavira gains 14 points with its wider camera and 3 points with the sensor-matched $42^\circ$ camera of Table~\ref{tab:ablation}, and in both configurations its SPL is about twice that of any adapted method.
The wider field of view thus accounts for a large part of the SR gap of \lavira but not for its efficiency.}

\noindent\textbf{The two paradigms are complementary.}
\higa is the strongest on category goals with 53.0\% ON but weak on image goals with 18.8\% IIN and the least efficient method with 7.1\% SPL, since its detector finds categories reliably but keypoint matching rarely confirms an instance, and its policy, trained for single-goal episodes, travels far.
\lavira is the strongest on instance goals with 41.4\% IIN and 40.9\% TN and the most efficient with 20.3\% SPL, since the VLM reasons over the reference image or description and \rev{shared progress analysis} avoids redundant travel, but its category search trails \higa by 14 points.
\rev{With the matched camera, \lavira still leads \higa on image goals, 27.8 against 18.8, and on SPL, 23.0 against 7.1.}
Neither paradigm dominates.

\begin{table}[t]
\caption{\textbf{Ablations} on the \bench validation split. \higa: assignment strategy with the navigation policy fixed. \lavira: coordination channels. Best within each group in \textbf{bold}.}
\label{tab:ablation}
\vspace{-6pt}
\centering
\small
\renewcommand{\arraystretch}{1.1}
\begin{tabular*}{0.97\linewidth}{@{\extracolsep{\fill}} l c c c c c @{}}
\toprule
\textbf{Variant} & \textbf{SR}$\uparrow$ & \textbf{SPL}$\uparrow$ & \textbf{ON}$\uparrow$ & \textbf{TN}$\uparrow$ & \textbf{IIN}$\uparrow$ \\
\midrule
\rowcolor{black!10}\multicolumn{6}{l}{\textbf{\higa}, assignment strategy} \\
random & 16.8 & 2.7 & 22.7 & 17.3 & 8.3 \\
rule-based & 21.6 & 4.3 & 29.5 & 19.7 & 11.3 \\
\higa-BC & 34.1 & 6.0 & 44.7 & \textbf{41.7} & 14.3 \\
\higa-DAgger & \textbf{37.9} & \textbf{7.1} & \textbf{53.0} & 40.2 & \textbf{18.8} \\
\midrule
\rowcolor{black!10}\multicolumn{6}{l}{\textbf{\lavira}, coordination channels} \\
Full & \textbf{41.0} & 20.3 & \textbf{38.6} & \textbf{40.9} & \textbf{41.4} \\
w/o SG & 33.2 & 21.6 & \textbf{38.6} & 26.8 & 33.8 \\
w/o SP & 37.3 & 17.2 & 35.6 & 36.2 & 39.8 \\
w/o SG + SP & 29.3 & 18.8 & 30.3 & 26.0 & 30.1 \\
\arrayrulecolor{black!35}\specialrule{0.3pt}{1.5pt}{1.5pt}\arrayrulecolor{black}
\rev{w/ \higa camera} & \rev{29.6} & \rev{\textbf{23.0}} & \rev{32.6} & \rev{27.6} & \rev{27.8} \\
\bottomrule
\end{tabular*}
\renewcommand{\arraystretch}{1.0}
\vspace{-8pt}
\end{table}

\subsection{Ablations}
\noindent\textbf{\higa.}
With the navigation policy fixed, Table~\ref{tab:ablation} compares four assignment strategies.
Random assigns goals uniformly at random, rule-based matches agents to detected candidates by travel cost, and \higa-BC and \higa-DAgger are trained by behavior cloning of the oracle assigner without and with three DAgger rounds.
Learning to assign is worth 16 points over the hand-designed rule, because the learned assigner acts on the heat maps \emph{before} a goal is detected, whereas the rule can only react to detections.
DAgger adds 3.8 points over behavior cloning, concentrated on ON and IIN.

\noindent\textbf{\lavira.}
Both coordination channels matter, and for different reasons.
In the variant without SG, the shared goal board is replaced by an explicit assignment that gives each agent one goal.
SR drops by 7.8 points, almost entirely on instance goals, with TN falling from 40.9 to 26.8 and IIN from 41.4 to 33.8, while SPL rises slightly from 20.3 to 21.6.
A fixed assignment makes paths direct but prevents agents from claiming goals they encounter opportunistically, which is how most instance goals are found.
In the variant without SP, teammates' \rev{progress analysis} is withheld, and SR drops by 3.7 points uniformly across modalities and SPL by 3.1 points, because agents that do not know where their teammates have been re-explore, and removing both channels costs 11.7 points.
\rev{The last row, separated by a rule, is the sensor-matched control in which \lavira uses the $42^\circ$ Stretch camera of \higa and the adapted methods instead of its $79^\circ$ camera. SR falls by 11.4 points across all modalities while SPL rises to 23.0, since a narrower view finds fewer goals but reaches those it finds with less detouring.}

\begin{figure}[t]
    \centering
    \includegraphics[width=0.97\linewidth]{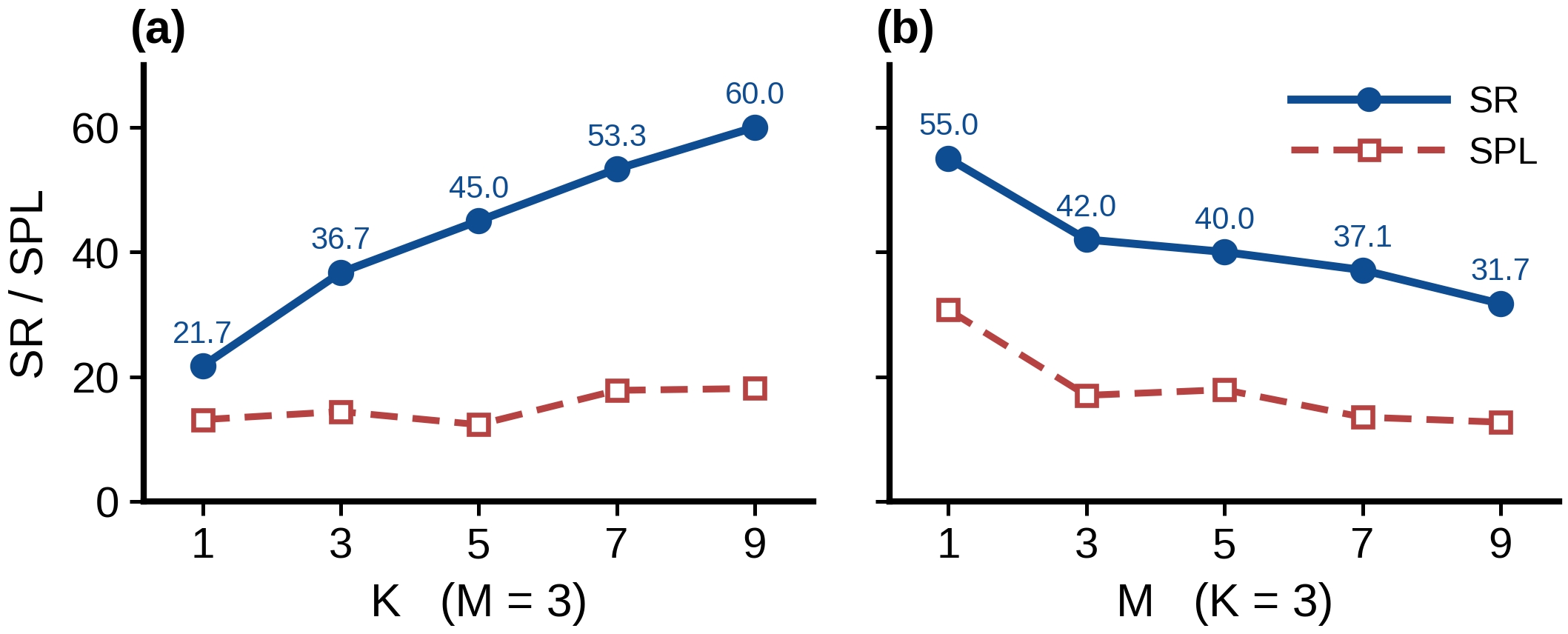}
    \vspace{-6pt}
    \caption{\textbf{Scaling with team size and goal count} for \lavira, with 200 episodes per point on validation scenes. (a) SR grows monotonically with $K$ at $M{=}3$ while SPL stays flat. (b) SR and SPL fall with $M$ at $K{=}3$.}
    \label{fig:scaling}
    \vspace{-8pt}
\end{figure}

\begin{figure*}[t]
    \centering
    \includegraphics[width=0.97\linewidth]{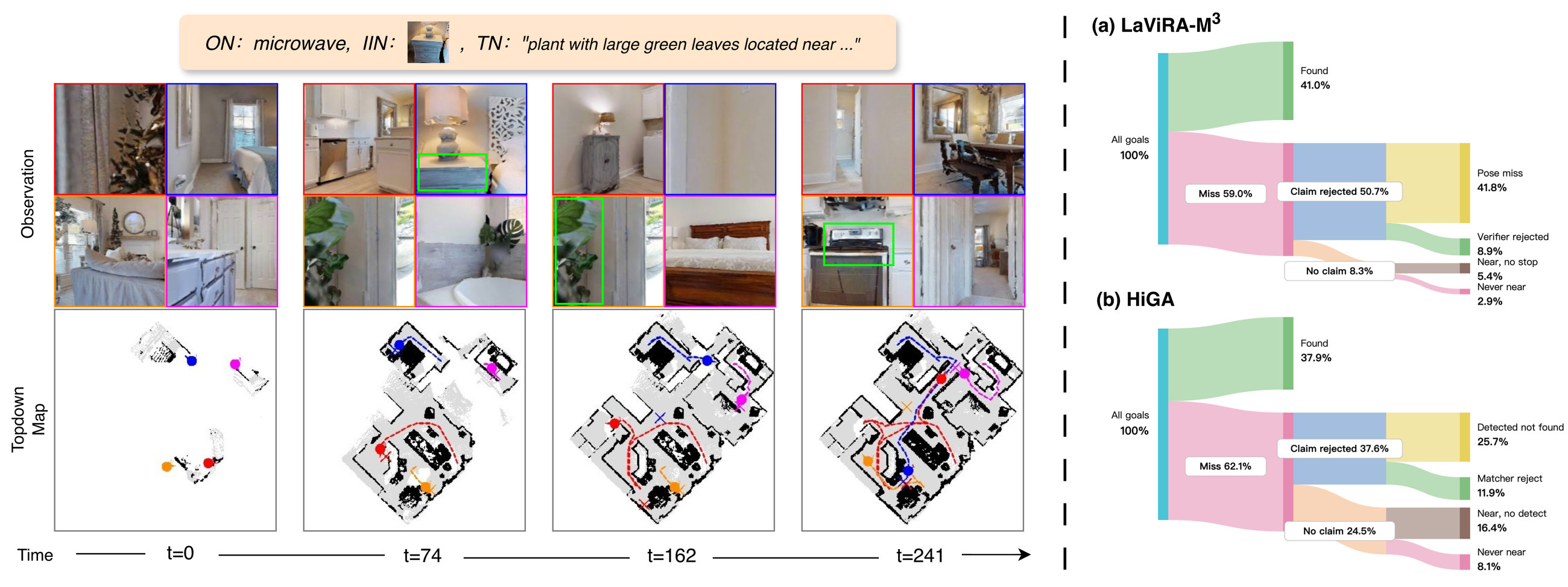}
    \vspace{-4pt}
    \caption{\textbf{Qualitative example and failure attribution.} Left, a $K{=}4$, $M{=}3$ episode of \lavira with one goal per modality, with agents in different colors. Right, where missed goals go, as a fraction of all goals, for (a) \lavira and (b) \higa. A missed goal is either \emph{claimed} and rejected by the protocol, or never claimed.}
    \label{fig:failure}
    \vspace{-8pt}
\end{figure*}

\begin{figure*}[t]
    \centering
    \includegraphics[width=0.97\linewidth]{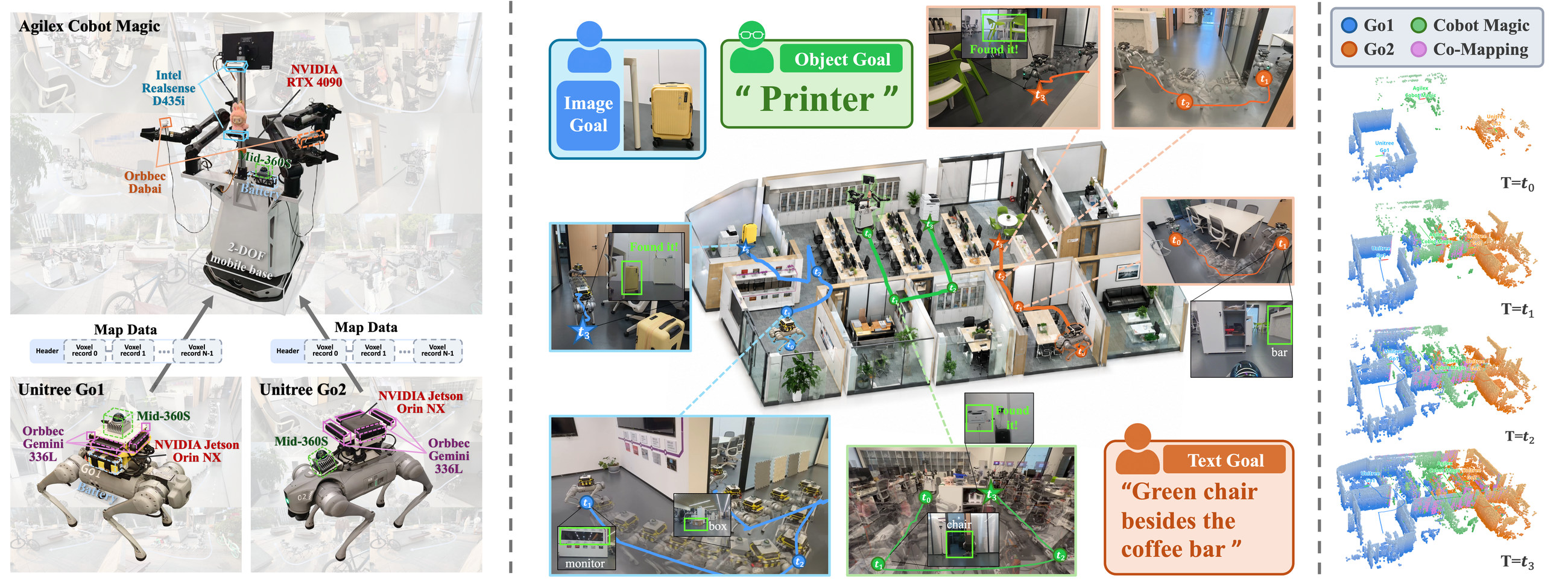}
    \vspace{-4pt}
    \caption{\rev{\textbf{Real-world deployment.} Left, the three platforms---an Agilex Cobot Magic, a Unitree Go1, and a Unitree Go2---exchanging the shared map as voxel records. Center, a run with three goals---an image of a yellow suitcase, the category ``printer'', and the description ``green chair besides the coffee bar''---with first-person views at $t_0$ to $t_3$ ending when each robot declares its goal found. Right, the shared map growing over time, colored by robot, with regions mapped by more than one robot in pink.}}
    \label{fig:realworld}
    \vspace{-8pt}
\end{figure*}

\subsection{Scaling with $K$ and $M$}
Fig.~\ref{fig:scaling} varies one axis while fixing the other, with 200 \lavira episodes per point on the validation scenes.
Adding agents helps, as SR at $M{=}3$ rises monotonically from 21.7\% with one agent to 60.0\% with nine.
\rev{SPL, however, stays at 12--18\%, which is the cost side of the same curve: additional agents buy coverage rather than a proportionally shorter division of labor, and current coordination does not yet exploit the parallelism that $d^\ast$ assumes.}
Adding goals hurts, as SR at $K{=}3$ falls from 55.0\% with one goal to 31.7\% with nine and SPL falls from 30.7\% to 12.7\%, since each additional goal competes for the same team budget and lengthens the makespan.

\subsection{Failure Analysis}
\label{sec:failure}
Fig.~\ref{fig:failure} attributes every missed goal to one cause, for both baselines, and the first split is whether the method ever acted on the goal.
\lavira misses 59.0\% of all goals, and for 50.7\% of all goals it acted and still missed.
In 41.8\% the agent declared \texttt{FOUND} but stood more than 1\,m from a valid viewpoint, which we call a pose miss, and in 8.9\% its own verification rejected the claim and it never returned.
Only 8.3\% of goals were never claimed, and only 2.9\% were never approached within 3\,m.
\rev{The pose misses are not one failure but three.
Only about a quarter were declared within 2\,m of a viewpoint, the case in which the agent saw the object and stopped early, which the 2\,m depth gate of Sec.~\ref{sec:lavira} permits.
About 30\% had passed within 1\,m of the goal earlier but declared elsewhere, a failure of choosing where to stop, and about 40\% were declared more than 5\,m away, that is, the agent misidentified an object or a room.
The looser gate therefore explains only part of the gap. Stop selection and identity verification at a distance are the larger parts, and all three are properties of the baseline rather than of the task.}
\higa misses 62.1\%, and for 37.6\% it acted and still missed.
In 25.7\% the detector fired but the approach did not end within 1\,m of the right instance, and in 11.9\%, all of them image goals, the keypoint matcher never confirmed the instance.
A further 16.4\% of goals were within 3\,m of an agent but never detected, and 8.1\% were never approached.
Perception rather than allocation limits the trained paradigm.
\rev{The two profiles are complementary and suggest pairing a VLM that recognizes instances with a learned policy that approaches, selects the stop point, and confirms identity at close range.}
\section{Real-World Deployment}
\label{sec:realworld}

To test whether the setting transfers beyond simulation, we deploy \lavira on a heterogeneous team of three robots in an office, an Agilex Cobot Magic wheeled platform, a Unitree Go1, and a Unitree Go2 quadruped, shown in Fig.~\ref{fig:realworld}.
\rev{Each robot carries an RGB-D camera, a Livox Mid-360 LiDAR for localization and mapping, and an onboard computer, an RTX 4090 on the Cobot Magic and a Jetson Orin NX on each quadruped, on which it runs its own LaViRA agent with a locally deployed Qwen3.8-27B~\cite{qwen38}, the successor of the Qwen3.5 model used in simulation, in 4-bit GGUF quantization, without any cloud API.
The robots exchange the shared map as voxel records over Wi-Fi, together with the goal board and their progress analysis, through the same channels as in simulation, and each executes the resulting waypoints with its own velocity controller, the only platform-specific component.}
\rev{In the run of Fig.~\ref{fig:realworld}, the robots spread out through the office, each claims the goal it encounters, and a goal found by one robot is removed from the others' lists, as also shown in the supplementary video.}

\section{Conclusion}
\label{sec:conclusion}

We introduced \bench, a benchmark in which a team of agents \rev{searches for} an unordered set of open-vocabulary goals given as categories, instance images, and descriptions, with a trained and a training-free baseline.
The results show a task with ample headroom at 41\% SR, two baselines with complementary strengths and \rev{distinct failures}, and a difficulty that grows with the number of goals and is only partly relieved by adding agents.

\noindent\textbf{Limitations.}
We assume a shared coordinate frame and lockstep execution, without communication cost or localization drift.
Text goals are written by a VLM and only spot-checked, results come from a single run, and the real-world deployment is qualitative.

\bibliographystyle{IEEEtran}
\bibliography{refs}

\end{document}